\documentclass[11pt]{article}
\usepackage{coling2020}
\usepackage{times}
\usepackage{url}
\usepackage{latexsym}

\usepackage{graphicx}
\usepackage{tabulary}
\usepackage{nameref}

\usepackage{booktabs}
\usepackage{multirow}
\usepackage{multicol}
\usepackage{xspace}
\usepackage{xcolor}
\usepackage[ruled, vlined, linesnumbered]{algorithm2e}

\newcommand{\relA}{EHR-RelA\xspace}
\newcommand{\relB}{EHR-RelB\xspace}

\colingfinalcopy 

\title{Biomedical Concept Relatedness -- A large EHR-based benchmark}
\author{Claudia Schulz \and Josh Levy-Kramer \and Camille Van Assel \and \\
\bf{Miklos Kepes} \and \bf{Nils Hammerla}\\
  Babylon Health \\
  London, SW3 3DD, UK \\
}
  
\begin{document}
\maketitle
\begin{abstract}
A promising application of AI to healthcare is the retrieval of information from electronic health records (EHRs), e.g.~to aid clinicians in finding relevant information for a consultation or to recruit suitable patients for a study.
This requires search capabilities far beyond simple string matching, including the retrieval of concepts (diagnoses, symptoms, medications, etc.) \emph{related} to the one in question. 
The suitability of AI methods for such applications is tested by predicting the relatedness of concepts with known relatedness scores.
However, all existing biomedical concept relatedness datasets are notoriously small and consist of hand-picked concept pairs.
We open-source a novel concept relatedness benchmark overcoming these issues: it is six times larger than existing datasets 
and concept pairs are chosen based on co-occurrence in EHRs, ensuring their relevance for the application of interest.
We present an in-depth analysis of our new dataset and compare it to existing ones, highlighting that it is not only larger but also complements existing datasets in terms of the types of concepts included.
Initial experiments with state-of-the-art embedding methods show that our dataset is a challenging new benchmark for testing concept relatedness models.
\end{abstract}

%
%
\blfootnote{
    %
    %
    %
    %
    \hspace{-0.65cm}  
    This work is licensed under a Creative Commons 
    Attribution 4.0 International Licence.
    Licence details:
    \url{http://creativecommons.org/licenses/by/4.0/}.
    %
    %
}

\section{Introduction}
\label{sec:introduction}

The adoption of electronic health records (EHRs) facilitates interoperability, meaning that more and more information from different sources is being stored about a patient.
This makes it increasingly challenging for doctors to efficiently filter a patient's record for relevant information during a consultation without missing anything.
This is particularly problematic since consultations are time-constrained. In the UK for example, general practitioner (GP) doctors usually have less than 10 minutes to consult a patient \cite{Flaxman2015,Salisbury2019}.

EHRs consist not only of free-text records but are furthermore tagged by doctors with medical concept codes.
This coding is aimed at standardising health records to enable, e.g., the seamless transfer of patient information between practices and the analysis of health data from different practices \cite{ehrBook,MorrisonEtAl2014}.
In addition, coded EHRs allow for the search of concept codes in an EHR.
However, retrieving not only the exact concept in question but also \emph{related} ones, as done by doctors when reading a patient's record, is less straight-forward.

For a patient with potential liver failure, related information  of interest to a doctor in the patient's history are for example `alcohol abuse', as it is a risk factor of liver failure, and `jaundice', a symptom of liver failure. 
Other risk factors, symptoms, treatments, conditions, or tests associated with liver failure would also be considered relevant. 

Concept representation models, such as embeddings or ontology-based methods \cite{McInnesPP2009,Pivovarov2012,Henry2018,SmalheiserCB2019,ParkKHL2019}, have been developed to tackle the task of identifying and retrieving related concepts.
These methods have the potential to aid doctors in finding related information in a patient's EHR, which can increase the quality of medical outcomes by improving efficiency, as consequently alleviating time pressure, and ensuring that doctors do not miss important information.
It is however unclear how well these methods would perform in real-world EHR concept retrieval settings as they have so far only been tested on very small datasets, as pointed out by \newcite{SchulzJ2020}.

We address this issue by constructing a novel open-source\footnote{\url{https://github.com/babylonhealth/EHR-Rel}} biomedical concept relatedness dataset consisting of 3630 concept pairs -- six times more than the largest existing dataset.
Instead of manually selecting and pairing concepts as done in previous work, our dataset is sampled from EHRs to ensure concepts are relevant for the EHR concept retrieval task.
The relatedness scores assigned to concept pairs in our dataset are of high quality, as shown by good inter-annotator agreement and reliability metrics.
A detailed analysis of the concepts in our novel dataset reveals a far larger coverage compared to existing datasets.
We furthermore report the results of initial experiments with state-of-the-art embeddings, illustrating that our dataset constitutes a challenging new benchmark.

\section{Related Work}
\label{sec:related-work}
Relatedness and similarity are not to be confused, even though embedding models are often tested on both types of relations \cite{Chiu2018,HenryMM2019,SchulzJ2020}.
Semantic similarity is a specific type of semantic relatedness
\cite{Pakhomov2010,Pakhomov2011}, meaning that similar concepts are generally related but not vice versa.
As an example, `liver failure' and 'alcohol abuse' are medically related but not semantically similar.
We are here concerned with \emph{relatedness}.

\newcite{Pedersen2007} hand-picked 120 pairs of medical concepts from UMLS \cite{Bodenreider_2004} 
that were expected to have a balanced distribution across four categories: closely related, somewhat related, somewhat unrelated and completely unrelated.
The pairs were then rated by 13 medical coders on a 1-10 relatedness scale. Coders were not given a definition of the scale and were instructed to use their intuition. Since the coders' agreement was low, 
the 29 concept pairs with highest agreement were chosen and annotated again by nine medical coders and three physicians as synonyms (4), related (3), marginally related (2), or unrelated (1), 
resulting in the \emph{MiniMayoSRS} dataset.

\newcite{Pakhomov2011} selected a subset of 101 pairs from the original 120, excluding duplicates and  ambiguous pairs, and analysed the coders' ratings in more detail. This subset is available as the \emph{MayoSRS} dataset. Based on their observations, they also proposed a framework for the future creation of concept relatedness datasets, which we closely follow.
MiniMayoSRS and MayoSRS both lack size and coverage. They are thus not suitable for testing concept relatedness models with the purpose of selecting the best one for real-world applications.

\newcite{Pakhomov2010} introduced the \emph{UMNSRS-Sim} 
and \emph{UMNSRS-Rel} datasets, consisting of manually chosen
pairs of UMLS concepts rated on a continuous scale of 0-1600 regarding their similarity and relatedness, respectively. The rating was performed on a touch screen, where the continuous scale corresponds to pixels on the screen. Raters had only 4 seconds to rate each pair and were not given any definition of the similarity/relatedness scale.
Out of 724 given concept pairs, four medical coders rated 566 of them regarding similarity and another four coders 587 of them regarding relatedness.

\newcite{hliaoutakis2005semantic} presented 36 pairs of MeSH terms with a similarity score of 0-1.
\newcite{Chiu2018} created the \emph{Bio-SimLex} and \emph{Bio-SimVerb} datasets of, respectively, 988 pairs of nouns and 1000 pairs of verbs that frequently occur in PubMed. 
Since both works involve \emph{similarity} rather than relatedness and \emph{terms} are not linked to concepts in any biomedical ontology, they are omitted from our comparison of existing datasets with our novel benchmark.

\section{A New Concept Relatedness Benchmark}

To enable the development of reliable models for searching EHRs and biomedical literature, appropriate benchmark datasets for testing are essential. Existing datasets (see Section~\ref{sec:related-work}) have various shortcomings, which we address in the construction of our novel benchmark:
1) \textbf{Size:} An appropriate test set needs to be of sufficient size to allow for the generalisability of performance results. Our novel benchmark is 6 times larger than the biggest existing dataset.\\
2) \textbf{Concept selection:} For existing datasets UMLS concepts were manually chosen, so it is unclear how relevant the chosen concepts are for an application such as EHR search. In contrast, we automatically retrieve frequently occurring concepts from EHRs.\\
3) \textbf{Annotation guidelines:} Rather than relying on the annotators' intuition as to what `relatedness' means, we follow the suggestion of \newcite{Pakhomov2011} to provide annotators with clear guidelines about the relatedness scale.\\
4) \textbf{Relatedness scale:} \newcite{Pakhomov2011} furthermore suggested to use a small scale. Our new benchmark has a 0-3 scale, fulfilling this requirement.

In the following, we describe the selection of concept pairs for our dataset and their annotation.
\subsection{Constructing medical concept pairs from IMRD}
\label{sec:construction-and-content}
In contrast to existing datasets, where concepts are either manually selected from UMLS/MeSH or sampled from PubMed and then paired, we directly sample concept pairs from EHR data. 
In particular, we use IQVIA Medical Research Data (IMRD) incorporating data from The Health Improvement Network (THIN, a Cegedim database), which consists of anonymised primary care EHRs, covering 5\% of the UK population.

A patient's consultation in IMRD may include concepts from the following categories: symptom, diagnosis, presenting complaint, examination, intervention, management, and administration. 
We here only consider concepts from the the first three categories as they are the most relevant to the purpose of EHR search to aid consultations.
For each patient in IMRD, we pair all distinct concepts (from the three categories) occurring in the patient's EHR, resulting in a total of 1,345,193 unique pairs made from 34,794 unique concepts.

The concepts in IMRD are given as Read Version 2 codes, a coding system that is almost exclusively used in the UK \cite{RobinsonEtAl2997}. To ensure international compatibility, we map all concepts in the extracted concept pairs to SNOMED-CT IDs \cite{Donnelly2006} using the mappings provided by NHS Digital\footnote{\url{https://isd.digital.nhs.uk/trud3/user/guest/group/0/pack/8/subpack/9/releases}}. 

Despite belonging to the symptom, diagnosis, or presenting complaint categories, some of the extracted concepts describe administrative or navigational rather than medical information, e.g.~``did not attend appointment'' or ``situation with explicit context''. 
Such concepts are manually flagged and filtered out along with all their descendants specified in SNOMED-CT.
The mapping and filtering results in 1,066,541 unique concept pairs made of 30,276 unique concepts represented by SNOMED-CT IDs.

\subsection{Annotation scale and setup}

Our five annotators are experienced doctors, registered and licensed with the General Medical Council (GMC).
To ensure that all annotators have the same understanding of relatedness, we define a relatedness scale of zero to three, as shown in Table \ref{tab:relatedness-scale}, based on detailed discussions with doctors. We also perform a small pre-annotation study with all annotators to train them in applying the relatedness scale and to discuss potential misunderstandings and difficulties.

\begin{table}[t]
\centering
\small
\begin{tabulary}{\linewidth}{|c|L|}
\hline
\textbf{Score} & \textbf{Definition}                                              \\ \hline                                                                 
\textbf{0}     & \textit{\textbf{Unrelated:}} completely unrelated concepts -- the concepts have nothing in common and no relationship links them                                                                      \\ \hline
\textbf{1}    & \textit{\textbf{Marginally related:}} there is a correlation between the concepts, but an established link might not exist                                                                             \\ \hline
\multirow{2}{*}{\textbf{2}}     & \textit{\textbf{Related:}} the concepts are strongly related medically, e.g.~one leads to the other (nausea leads to vomiting), or the concepts have an established link (obesity and ischemic heart disease) \\ \hline
\multirow{2}{*}{\textbf{3}}     & \textit{\textbf{Extremely related:}} the concepts always occur together medically, or one cannot happen without the other (alcoholic liver disease and liver cirrhosis)                                      \\ \hline
\end{tabulary}
\caption{{Relatedness scale used for annotations.}}
\label{tab:relatedness-scale}
\end{table}

\paragraph{\relA:} We first \emph{randomly} select 120 pairs from the list of concept pairs for annotation by all five annotators.
Our analysis of these annotations (details are discussed in Section~\ref{sec:analysis}) shows that the distribution of the relatedness scores is highly skewed towards non-related concepts. 

\paragraph{\relB:} To create a more balanced dataset, we sample concept pairs based on the assumption that concept pairs occurring frequently in EHRs are more likely to be related.
The 1,066,541 unique concept pairs are therefore sorted by their number of occurrences in descending order. 
The pairs are then filtered so that only the top six occurrences of each concept are included to ensure a higher coverage of unique concepts in our dataset. We then choose the top 4,000 concept pairs, which include 2479 unique concepts. 
Since our analysis of the preliminary \relA annotations show good annotator agreement (see Section~\ref{sec:analysis}), each of the 4,000 concept pairs is annotated by only three annotators to save resources (different concept pairs are annotated by a different subset of three out of the five annotators).

\section{Dataset Analysis}
\label{sec:analysis}

Some concept pairs in \relA and \relB were not rated as the meaning of some concepts was unclear. Excluding these pairs, \relA consists of 111 concept pairs and \relB of 3630.

\subsection{Annotation quality and reliability}
To assess the quality of our annotated datasets as well as the difficulty of the task, we analyse the annotators' agreement.
We closely follow the methodology set out by \newcite{Pakhomov2011}, considering 1) inter-annotator agreement, measured as pairwise coefficients between each of the annotators, and 2) multi-rater reliability, measured as summary statistics of all annotators together.

\subsubsection*{Inter-annotator agreement}
Pairwise agreement measures are useful in identifying single annotators with low performance as well as disagreements between pairs of annotators.
Following \newcite{Pakhomov2011}, we use three measures: Spearman's $\rho$ (correlation), Cohen's $\kappa$ and Krippendorff's $\alpha$.

The agreement between all annotators in terms of Krippendorff's $\alpha$ is 0.64 for \relA annotation and 0.59 for \relB.
The higher agreement on the 111 \relA concept pairs as compared to the \relB annotation of 3630 concept pairs can be attributed to the fact that the \relA annotation was highly skewed towards unrelated concept pairs, as will be shown in Section~\ref{sec:analysis_distribution}.

The pairwise Krippendorff's $\alpha$ agreement is presented in Table~\ref{tab:pairwise_alpha}. We omit the pairwise $\kappa$ and $\rho$ scores for space reasons and as they follow the trends of the pairwise $\alpha$ measure and only report the averaged $\kappa$ and $\rho$ scores for each annotator.
The table shows that overall there is a satisfactory agreement between all annotators.
The pairwise agreement scores reveal that annotator E agrees least with the other annotators. However, the agreement is still moderate, so we include annotator E's annotations in our dataset.
Since we publish not only the average relatedness score but also each annotator's individual annotations, future studies are free to exclude concept pairs with high disagreement.

\begin{table}
\centering
\small
\begin{tabular}{ll cccccccc cc}
\toprule
                  && \multicolumn{2}{c}{\textbf{Ann. A}} & \multicolumn{2}{c}{\textbf{Ann. B}} & \multicolumn{2}{c}{\textbf{Ann. C}} & \multicolumn{2}{c}{\textbf{Ann. D}} & \multicolumn{2}{c}{\textbf{Ann. E}}\\ 
                  \cmidrule(lr){3-4} \cmidrule(lr){5-6} \cmidrule(lr){7-8} \cmidrule(lr){9-10} \cmidrule(lr){11-12}
                  && RelA & RelB & RelA & RelB& RelA & RelB& RelA & RelB& RelA & RelB \\
                  \midrule
\parbox[t]{4mm}{\multirow{4}{*}{\rotatebox[origin=c]{90}{\textbf{pairw. $\alpha$}}}}
&\textbf{Ann. B} & 0.77 & 0.57     \\
&\textbf{Ann. C} & 0.69 & 0.58      & 0.65 & 0.64       \\
&\textbf{Ann. D} & 0.70 & 0.62      & 0.73 & 0.63      & 0.68 & 0.66      \\
&\textbf{Ann. E} & 0.52 & 0.49      & 0.40 & 0.57      & 0.57 & 0.54      & 0.50 & 0.55  \\ \midrule
&\textbf{Average $\alpha$} & \textbf{0.67 }  & \textbf{0.57}    & \textbf{0.64 }  & \textbf{0.60}  & \textbf{0.65 }  &  \textbf{0.60} & \textbf{0.65 }   & \textbf{0.61}  & \textbf{0.50 } & \textbf{0.54} \\
& Average $\kappa$ & 0.74 & 0.56 & 0.70 & 0.60 & 0.72 & 0.61 & 0.69 & 0.62 & 0.61 & 0.54\\
& Average $\rho$   & 0.70 & 0.58 & 0.69 & 0.62 & 0.67 & 0.63 & 0.68 & 0.63 & 0.60 & 0.55  \\
\bottomrule
\end{tabular}
\caption{{Pairwise Krippendorff's $\alpha$ and average $\alpha$, Cohen's $\kappa$, and Spearman's $\rho$ for each Ann(otator).}}
\label{tab:pairwise_alpha}
\end{table}

\subsubsection*{Rater reliability}
To assess the reliability of annotations, we follow \newcite{Pakhomov2011} in using Kendall's coefficient of concordance (Kendall's W) and the Intra-class Correlation Coefficients ICC(C,1) and ICC(C,k). 
\newcite{Mcgraw96} define 10 types of Intra-class Correlation Coefficient (ICC) which depend on the use. As \newcite{Pakhomov2011}, we select ICC(C,1) and ICC(C,k), because: 1) they consider annotators as representative of a larger population of similar annotators, in our case doctors,
and 2) they measure consistency instead of the absolute agreement, i.e.~systematic errors of an annotator are cancelled out.
ICC(C,1) measures the reliability of a single rater selected from the larger rater population, whereas
ICC(C,k) measures the reliability of an average of multiple raters from the larger rater population.

As shown in Table~\ref{tab:comparison}, the annotation reliability is good to excellent \cite{Cicchetti94}. As can be expected from the inter-annotator agreement analysis, the reliability on \relA is higher.
We also observe that ICC(C,1) is lower than ICC(C,k), indicating that the average annotation score is more reliable than a single annotator's scores.

In comparison to existing datasets, Table~\ref{tab:comparison} shows that the inter-annotator agreement on our datasets in terms of average Spearman's $\rho$ is higher than for the MayoSRS dataset. Note that the agreement for MiniMayoSRS is very high since only high-agreement concept pairs were chosen (see Section~\ref{sec:related-work}). 
Furthermore, the reliability of each of our individual annotators, as indicated by ICC(C,1), is higher than for MayoSRS. The higher average reliability (given by ICC(C,k)) for MayoSRS can be attributed to the much higher number of annotators.
For UMNSRS, only one reliability metric is given, which is lower than for our datasets.
The comparison shows that the \emph{quality} of annotations in our datasets is at least as high, if not higher, than for existing datasets.

From here onward, we consider the average of all annotations for a concept pair as its relatedness score.

\begin{table}[t]
\small\centering
\begin{tabular}{l rr ccccccc}
\toprule
\textbf{Dataset}    & \textbf{Pairs} & \textbf{Concepts} & \textbf{Scale} &  \textbf{Annotators} & \textbf{Avg $\rho$} & \textbf{ICC(C,1)} & \textbf{ICC(C,k)} & \textbf{Kendall's W}\\ \midrule
\multirow{2}{*}{\textbf{MiniMayoSRS}} & \multirow{2}{*}{29}  & \multirow{2}{*}{57}  &  \multirow{2}{*}{1-4}  &  3 physicians & 0.68$^{\dagger}$  & \multirow{2}{*}{-\phantom{*}}    & \multirow{2}{*}{-} & \multirow{2}{*}{-} \\
& & & & 9 medical coders   & 0.78$^{\dagger}$  \\
\textbf{MayoSRS}     & 101  & 182 & 1-10 & 13 coders & 0.53\phantom{$^{\dagger}$}   & 0.50\phantom{*}    & 0.93  & 0.57 \\
\textbf{UMNSRS-Rel} & 587  & 386   &  0-1600 & 4 medical residents & -\phantom{$^{\dagger}$} & 0.50*  & - & -\\ \midrule
\textbf{\relA}  & 111   & 161  & 0-3  & 5 doctors & 0.67\phantom{$^{\dagger}$}  & 0.72\phantom{*}  & 0.93  & 0.73 \\
\textbf{\relB}  & 3,630  & 2,263 & 0-3 & 3 doctors & 0.59\phantom{$^{\dagger}$} & 0.60\phantom{*}  & 0.81 & 0.73 \\
\bottomrule
\end{tabular}
\caption{\relB compared to existing datasets. *Unclear which ICC(C, $\cdot$) the authors used, so we assume ICC(C,1). $^{\dagger}$Unclear which correlation the authors used, so we assume Spearman's $\rho$.}
\label{tab:comparison}
\end{table}

\subsection{Distribution of relatedness scores}
\label{sec:analysis_distribution}
\begin{figure}[ht]
\centering
  \includegraphics[width=0.39\textwidth]{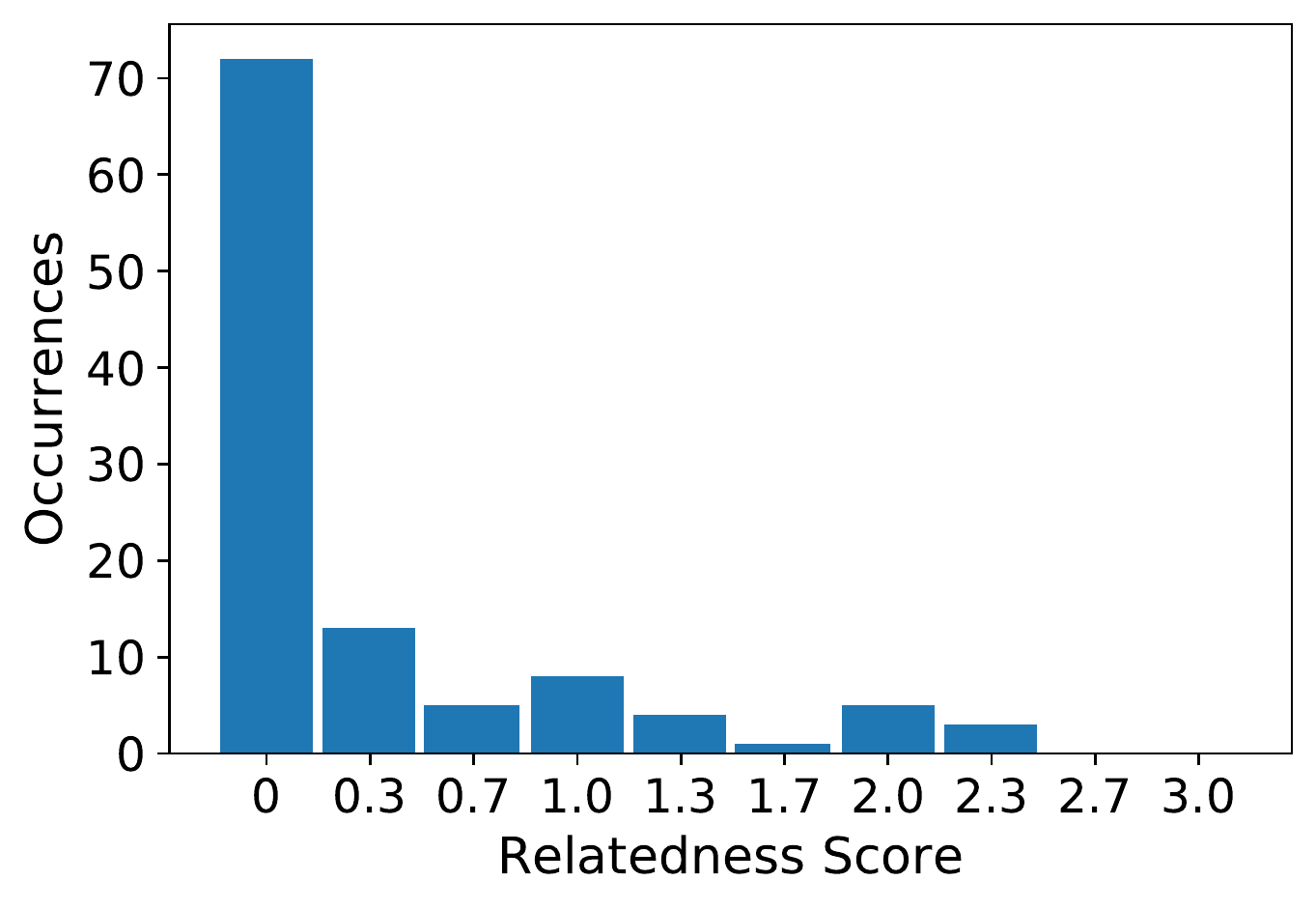} \hspace{0.6cm}
  \includegraphics[width=0.39\textwidth]{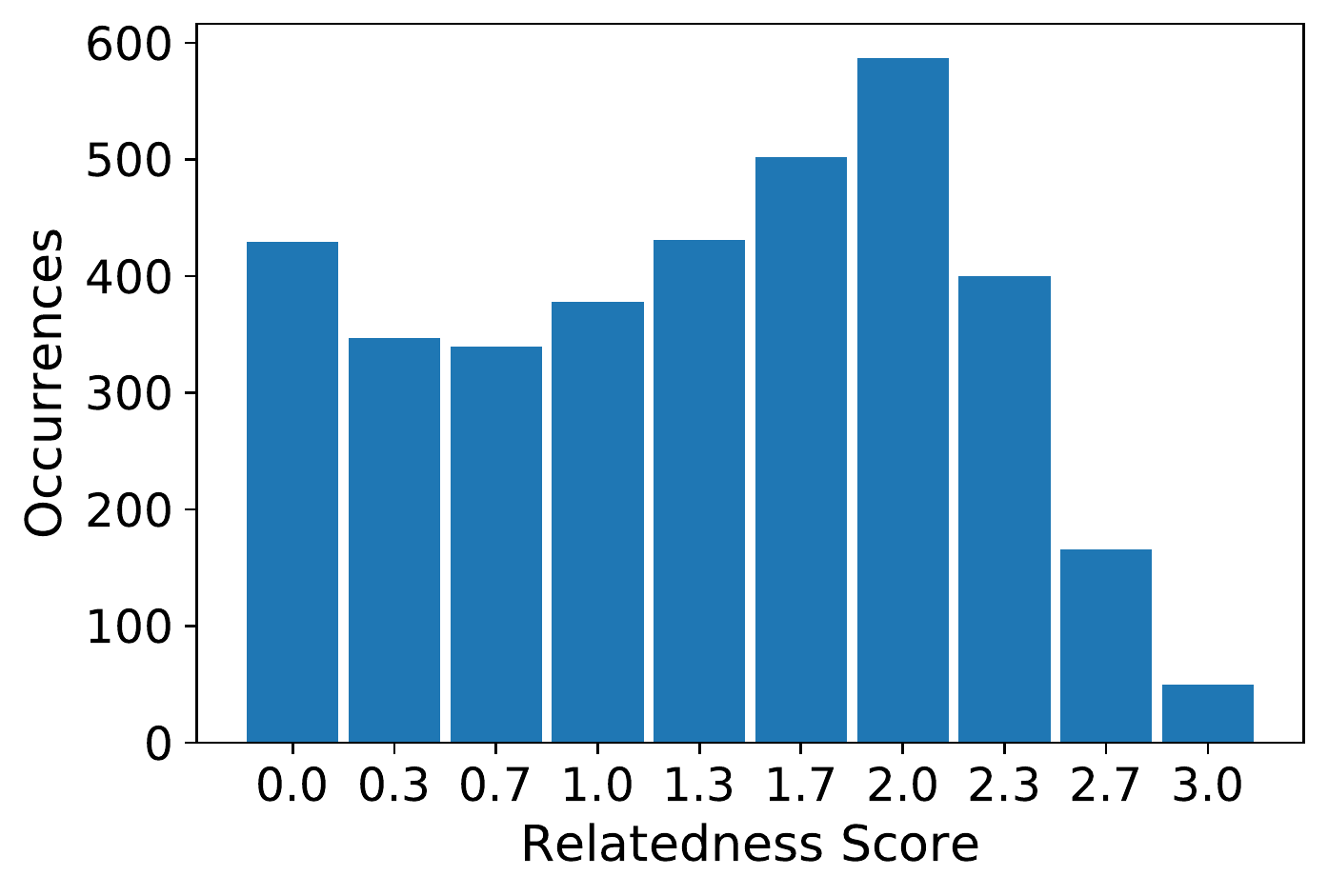}
  \caption{Distribution of relatedness scores in \relA (left) and \relB (right).}
  \label{fig:distribution}
\end{figure}

Figure \ref{fig:distribution} illustrates the distribution of relatedness scores in the two datasets. \relA is highly skewed towards `unrelated' pairs of concepts. This can be attributed to the random selection of co-occurring concept pairs.
In contrast, \relB was constructed by choosing the most frequently co-occurring pairs of concepts, leading to a balanced distribution of relatedness scores.

This is particularly interesting as \newcite{Pakhomov2011} found that hand-picking concept pairs to create balanced dataset is highly challenging. As we show, concepts frequently co-occurring in EHRs \emph{naturally} result in a balanced dataset.

Due to the small size and skewness of \relA, we consider and recommend only \relB as a new benchmark dataset. The analyses and experiments in the following sections therefore investigate \relB only.

\section{Concept Coverage}
\label{sec:coverage}
Clearly our new \relB dataset is larger than existing ones in terms of number of concept pairs and unique concepts.
In this section, we further investigate the \emph{types} of concepts in \relB compared to existing datasets. Since UMNSRS-Rel and UMNSRS-Sim consist of nearly the same concept pairs, their concept coverage is very similar. We thus only present results for the relatedness dataset UMNSRS-Rel. 

\subsection{Mapping SNOMED IDs to UMLS CUIs (Concept Unique Identifiers)}
The (Mini)MayoSRS as well as the UMNSRS-Rel datasets were constructed in terms of UMLS concepts \cite{Bodenreider_2004}. In contrast, our new dataset is made of SNOMED concepts.
To compare existing datasets with \relB, we thus map all SNOMED IDs in our new benchmark to UMLS CUIs, 
as detailed in Algorithm~\ref{alg:snomed2umls}. 
To get all CUIs associated with a SNOMED code (line~\ref{line:get_cuis}) and to find preferred SNOMED terms for a CUI (line~\ref{line:pref_snomed}), the
UMLS REST API\footnote{\url{https://documentation.uts.nlm.nih.gov/rest/home.html}} is used.

Since the SNOMED IDs in \relB are obtained from Read codes, some of them are not contained in the SNOMED-CT International version, as they are from the SNOMED-CT United Kingdom release, which is not included in UMLS.
Therefore, some SNOMED IDs in \relB cannot be mapped to a UMLS CUI.
The mapping results in 3225 pairs of UMLS concepts (out of the 3630 SNOMED pairs).

Note that expressing our new benchmark dataset in terms of UMLS CUIs is not only useful for the comparison with existing dataset, but also allows for the application of CUI embedding models  \cite{HenryMM2019,ParkKHL2019,Henry2018} for predicting concept relatedness.

\begin{algorithm}[t]
\small
\SetAlgoLined
\newcommand\mycommfont[1]{\footnotesize\ttfamily\textcolor{blue}{#1}}
\SetCommentSty{mycommfont}
\KwIn{$ids$ \tcc*[f]{list of SNOMED IDs}}
\KwOut{$map$ \tcc*[f]{dictionary of ID-CUI pairs}}
 \ForEach{$id \in ids$}{
   $cuis = get\_cuis(id)$ \tcc*[f]{all CUIs that $id$ is associated with} \label{line:get_cuis}\\
   \uIf(\tcc*[f]{filter CUIs to representative ones}){$len(cuis) > 1$}{
    $representative\_cuis = []$\\
    \ForEach{$cui \in cuis$}{
     $ps = get\_preferred\_snomed(cui)$ \tcc*[f]{all preferred SNOMED terms of $cui$}\label{line:pref_snomed}\\
     \uIf(\tcc*[f]{$id$ is a preferred term of $cui$}){$id \in ps$}{
      $representative\_cuis.append(cui)$ \tcc*[f]{thus $cui$ represents $id$}
      }
     }
     $cuis= representative\_cuis$
    }
    \uIf(\tcc*[f]{only consider unambiguous mappings}){$len(cuis) == 1$}{
    $map[id] = cuis[0]$
    }
    \uElse{$map[id] = None$}
 }
\caption{SNOMED to UMLS}
\label{alg:snomed2umls}
\end{algorithm}

\subsection{Semantic types}
\newcite{Pakhomov2010} constructed their concept pairs in the UMNSRS datasets by choosing concepts with semantic type `drug', `disorder', and `symptom' and combining them so as to obtain a balanced amount of semantic type combinations.
We chose concepts tagged as `presenting complaint', `diagnosis', or `symptom' in the EHRs, but did not use these tags to inform the creation of concept pairs. We thus analyse the semantic types of all UMLS CUIs in \relB as well as in existing datasets.

Many CUIs have more than one semantic type. Since UMLS semantic types are organised in a hierarchical \emph{semantic network}, we determine the most specific common ancestor of a CUI's semantic types and choose this to be its unique semantic type. Again, we make use of the UMLS REST API.

\subsubsection*{Distribution of semantic type combinations}
\begin{figure}[t]
  \centering
  \includegraphics[trim= 0 60 0 0, clip, width=0.75\textwidth]{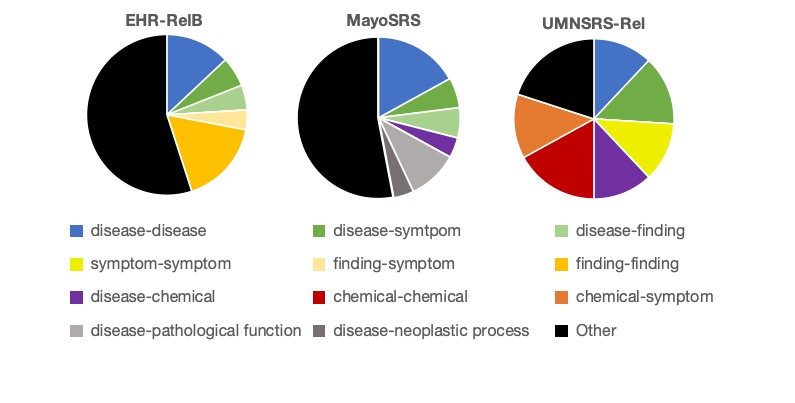}  
  \caption{Distribution of semantic type combinations in our benchmark and existing datasets.}
  \label{fig:sem_types}
\end{figure}

Figure~\ref{fig:sem_types} shows the distribution of the most common semantic type combinations (those making up more than 3\% of a dataset) in \relB compared to the existing MayoSRS and UMNSRS-Rel datasets.
Note that in UMLS the semantic type `symptom' is a specific type of `finding', so combinations involving these two semantic types are similar (e.g.~finding-symptom is similar to symptom-symptom).

The most frequently occurring semantic types in \relB are `symptom' and `disease', matching the three concept tags used to filter the IMRD EHRs.
21\% of concept pairs in \relB are of types finding-finding (or the similar finding-symptom), 13\% are disease-disease, and 11\% combine the two semantic types.
The rest of the concept pairs belong to one of 172 less frequent semantic type combinations.

As expected, UMNSRS-Rel exhibits an even distribution of semantic type combinations, in particular of `disease', `symptom', and `chemical'. Interestingly, none of the concepts in UMNSRS-Rel are of semantic type `clinical drug' (and `chemical' and `clinical drug' are only vaguely related as descendants of `physical object').
In contrast to \relB and UMNSRS-Rel, less than 4\% of concept pairs in MayoSRS are of type symptom-symptom (or similar combinations with `finding') and are thus not represented in Figure~\ref{fig:sem_types}. MayoSRS also has 10\% of concept pairs belonging to types disease-pathological function and 4\% to disease-neoplastic process, which occur much less frequently in the other datasets.
Note however that 10\% of MayoSRS constitutes only 10 concept pairs, whereas 10\% of \relB is 363 concept pairs.
 MiniMayoSRS consists of only 29 concept pairs, but includes 19 different semantic type combinations, so nearly all combinations occur only once. It is thus not included in Figure~\ref{fig:sem_types}.

There are 28 semantic type combinations in MayoSRS and UMNSRS-Rel that do not occur in \relB,  
13 of these contain a `chemical' semantic type, which is not represented in \relB.
Our new benchmark \relB comprises 148 combinations of semantic types not present in the existing datasets.

Our analysis shows 
1) that the distribution of most frequently co-occurring semantic types in EHRs, as given in \relB, is similar to that of the manually constructed existing datasets, and
2) that our new \relB benchmark \emph{complements} semantic type combinations in existing datasets.

\subsubsection*{Semantic type combinations versus relatedness scores}
Having analysed the distribution of semantic type combinations, we investigate whether any of the datasets has a bias of relatedness scores for the different semantic type combinations.
In other words, is the semantic type combination a good predictor of concept relatedness?
To answer this question, we compute the median relatedness score for each semantic type combination in a dataset. This is used as a baseline, predicting for each concept pair the median score of its semantic type combination.
The performance of this baseline is evaluated in terms of Spearman's correlation on concept pairs with a semantic type combination occurring more than once.

\begin{table}[b]
    \centering
    \small
    \begin{tabular}{l c c c}
    &  \textbf{MayoSRS} & \textbf{UMNSRS-Rel} & \textbf{\relB} \\ \midrule
    \textbf{Spearman's Correlation} & 0.46 & 0.23 & 0.33\\
    \end{tabular}
    \caption{Performance of the semantic type baselines for each dataset.}
    \label{tab:sem_type_baseline}
\end{table}

Table~\ref{tab:sem_type_baseline} shows that the semantic type combination of a concept pair is not a good predictor of relatedness in our new benchmark dataset or UMNSRS-Rel. This indicates that the relatedness scores in the datasets are \emph{not biased} by semantic types.
Note that this is the case despite the fact that the baselines are ``trained'' on the same data used for testing.
The higher correlation for the MayoSRS dataset can be attributed to its small size: concept pairs with a semantic type combination that occurs only two or three times, which is the case for many concept pairs in the MayoSRS dataset, are likely to have a more accurate median prediction than concept pairs belonging to a high frequency combination.
We omit the MiniMayoSRS dataset as its small size does not allow for a meaningful comparison.

\subsection{Concept specificity}
The previous section showed that our new \relB benchmark complements existing datasets in terms of semantic types of concepts.
Another interesting aspect of concepts is their specificity, i.e.~whether they are very general or specific concepts.
Given a hierarchical organisation of concepts, specificity can be defined in a straight-forward way in terms of a concept's shortest path from the root.
Since UMLS has no hierarchy of its own, we choose the SNOMED-CT hierarchy to measure specificity. 

\paragraph{UMLS CUIs to SNOMED IDs} To compare specificity in \relB with existing datasets, we map the UMLS CUIs in the existing datasets to SNOMED IDs.
As in Algorithm~\ref{alg:snomed2umls} line~\ref{line:pref_snomed}, all SNOMED IDs whose preferred term is associated with the CUI in question are obtained. If there are no such SNOMED IDs, the CUI's name as given in the dataset is used to search for SNOMED IDs.
We thus obtain a list of SNOMED IDs for each CUI. The specificity of a concept is then computed as the shortest path of any SNOMED ID in the list.
Note that for some CUIs in the UMNSRS-Rel dataset, it is not possible to find a matching SNOMED ID. 
We thus had to exclude 19 concept pairs from the analysis.

\begin{figure}
\centering
  \includegraphics[width=0.33\textwidth]{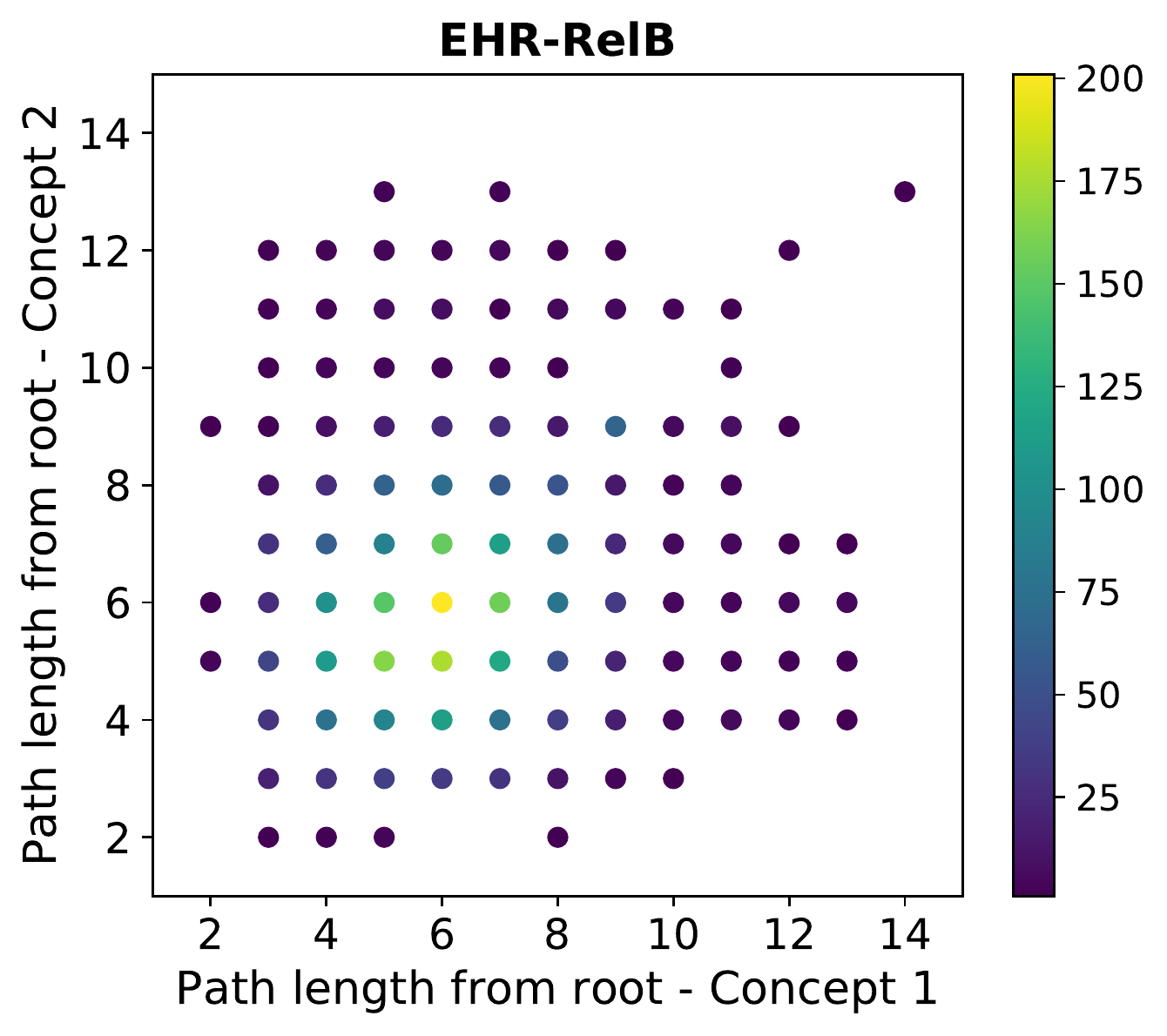}
  \includegraphics[trim= 25 0 0 0, clip, width=0.30\textwidth]{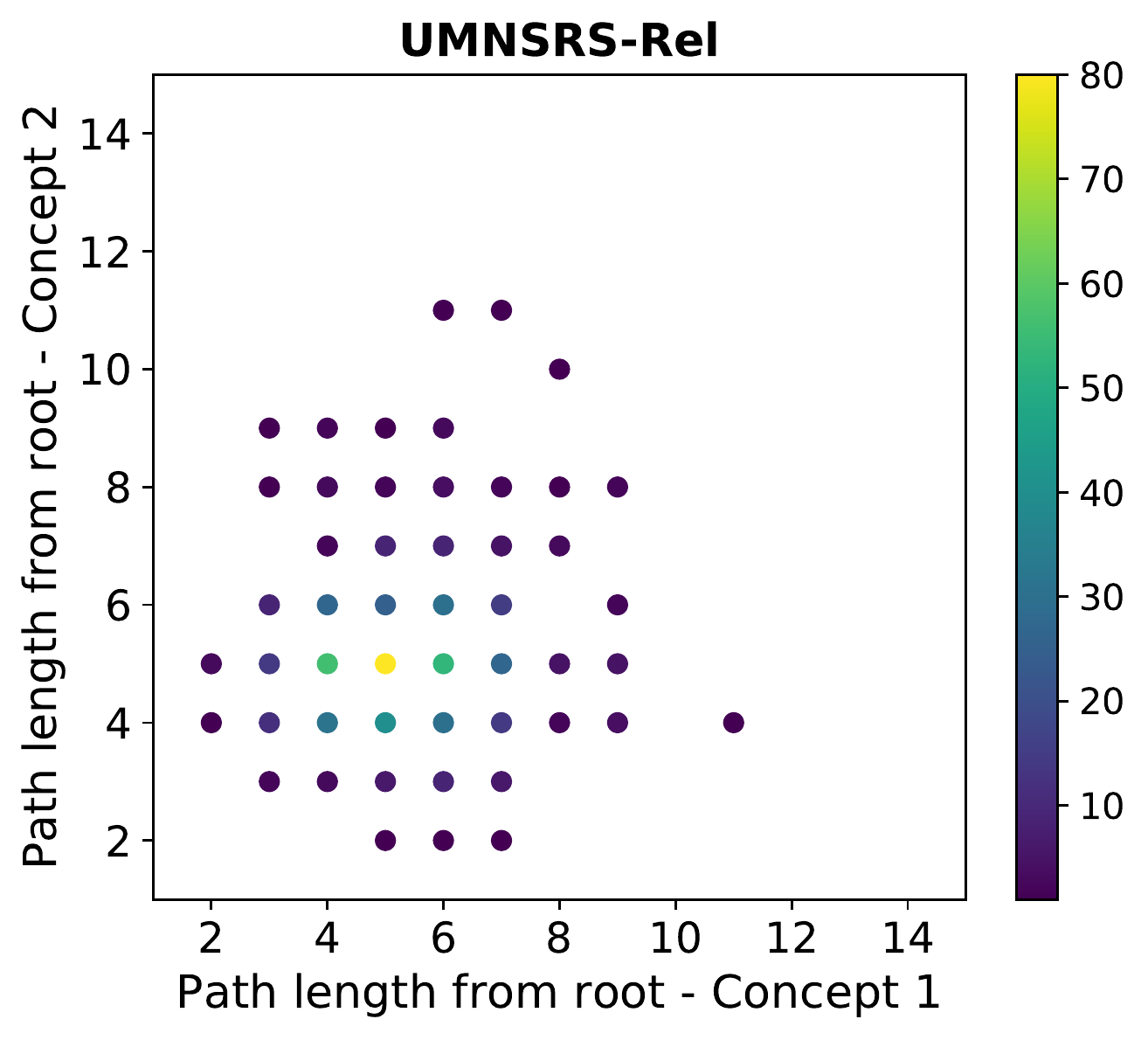}
  \includegraphics[trim= 25 0 0 0, clip, width=0.30\textwidth]{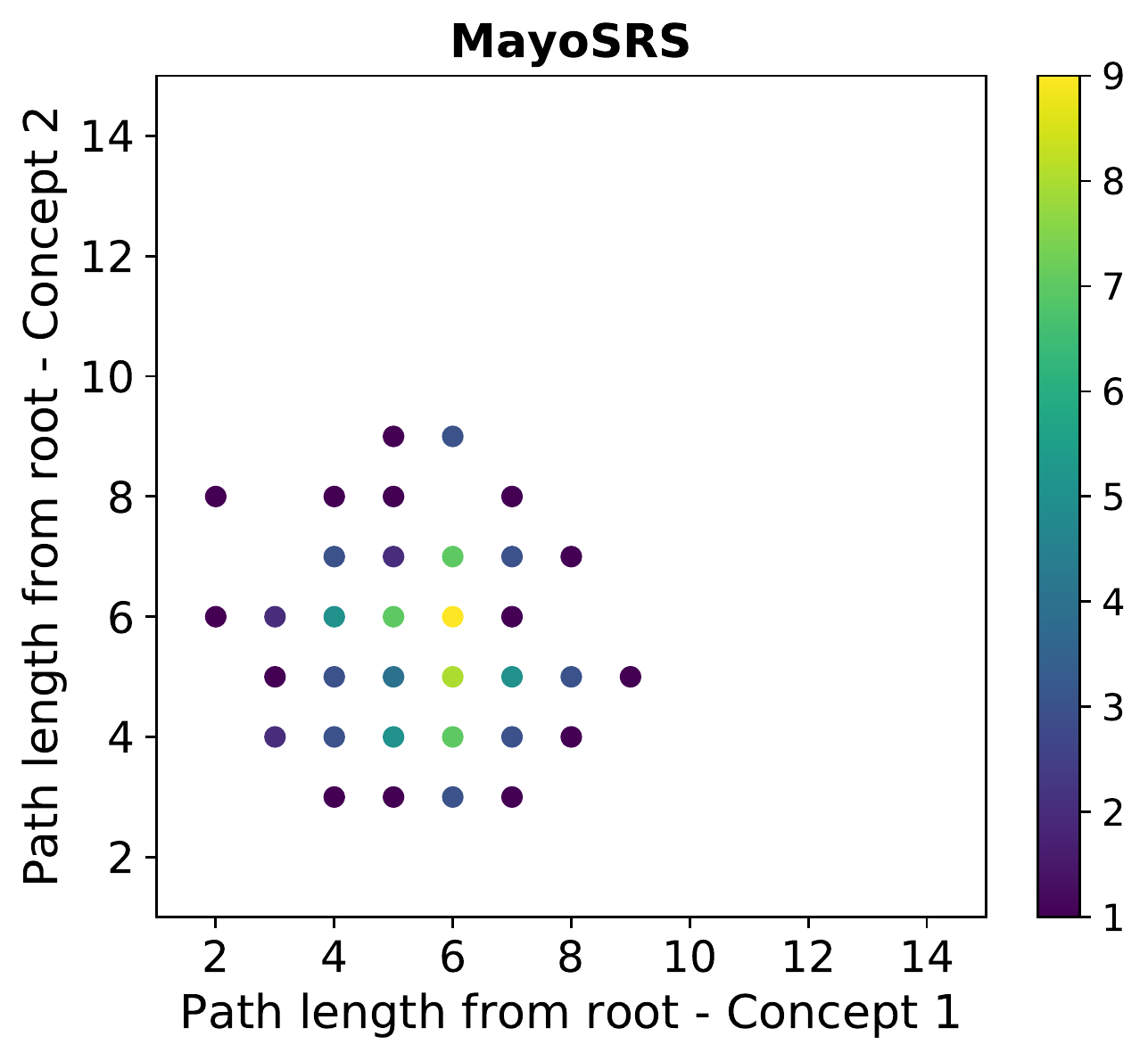}
  \caption{Distribution of concept specificity combinations in our new benchmark and existing datasets.}
  \label{fig:depth}
\end{figure}

Figure~\ref{fig:depth} illustrates the distribution of concept specificity combinations in \relB compared to UMNSRS-Rel and MayoSRS. 
We observe that concepts in existing datasets do not go beyond a specificity of 11, whereas our new benchmark \relB contains concepts with a maximum specificity of 14.
Furthermore, \relB also covers more general concepts, where the combination of most general concepts have specificity two and three.
The most frequent combination in \relB is of concepts with specificity six and six, which is the same as in MayoSRS.
In UMNSRS-Rel, the most frequent combination is of slightly more general concepts with a specificity of five and five.

Similar to the semantic type analysis, this evaluation shows that our new \relB benchmark \emph{goes beyond} existing datasets in terms of concept coverage as it adds more specific concepts while also containing very general ones. 
It also shows that existing datasets do not cover the breadth of concepts frequently occurring in EHRs. It is thus questionable how well the performance of concept relatedness models tested on existing datasets would generalise to real-world EHR concept retrieval. 

\section{Experiments with SOTA Embeddings}

As an initial experimental evaluation on our dataset, we evaluate the 13 state-of-the-art open-source biomedical word embeddings tested by \newcite{SchulzJ2020} on existing datasets: PMC, PM, PP, and PPW by \newcite{PyysaloEtAl2013}, ASQ by \newcite{KosmopoulosAP2015}, LTL2 and LTL30 by \newcite{ChiuEtAl2016}, AUEB2 (200) and AUEB4 (400) by \newcite{McdonaldEtAl2018}, (MeSH) extr and intr by \newcite{ZhangEtAl2019}, and MIM(IC) and its M(odel) version by \newcite{ChenPL2019}.
We do not consider the sentence embeddings tested by \newcite{SchulzJ2020} as they showed poor performance on existing datasets.

\begin{table}[t]
    \small
    \centering
    \begin{tabular}{l ccccccccccccc}
    \toprule
     \textbf{Sim.} &  PMC & PM & PP & PPW & ASQ & LTL2 & LTL30 & AUEB2 & AUEB4 & extr & intr &  MIM & MIM M \\ \midrule
     fJ & 0.43 & 0.46 & 0.45 & 0.44 & 0.48 & 0.44 & \textbf{0.49} & 0.46 & 0.46 & 0.48& 0.47 & 0.43 & 0.43\\
     cos & 0.40 & 0.44 & 0.42 & 0.41 & 0.47 & 0.36 & 0.41 & 0.40 & 0.40 & 0.35 & 0.37 & 0.33 & 0.33\\
    \bottomrule
    \end{tabular}
    \caption{Spearman's correlation for \relB with fuzzy Jaccard (fJ) or average cosine (cos) used to compute similarity for each embedding.}
    \label{tab:embeddings}
\end{table}

Table~\ref{tab:embeddings} shows the performance of each embedding on \relB in terms of Spearman's correlation.
Note that the performance is computed on a subset of 3350 out of the total 3630 concept pairs which could be embedded by all embeddings.
For each embedding we use both fuzzy Jaccard similarity \cite{ZhelezniakEtAl2019-fuzzyJaccard} and the standard average cosine as a similarity measure between vectors.
We observe that using fuzzy Jaccard similarity yields consistently higher performance for all embeddings.
LTL30 has the highest performance with a correlation of 0.49. Furthermore, it \emph{significantly outperforms}\footnote{We follow significance analysis as outlined by \newcite{SchulzJ2020}.} all but two embeddings (it does not outperform ASQ or extr). 
\newcite{SchulzJ2020} showed that most existing datasets are too small to observe significant difference between embeddings. Our results demonstrate that \relB is a promising new benchmark large enough to observe significant performance differences.

Compared to existing datasets, the performance of embeddings on \relB is lower. 
The best of the 13 embeddings on MayoSRS yields a Spearman's correlation of 0.57 and on UMNSRS-Rel 0.59 \cite{SchulzJ2020}.
To investigate possible reasons for the lower performance, 
we measure the performance of embeddings on a further subset of 2978 concept pairs, excluding concept pairs with high disagreement between annotators (3 different scores assigned). However, this only marginally improves performance, 
indicating that low agreement concept pairs are not a source of the lower model performance.
A possible explanation for the lower performance is that \relB consists of 89\% multi-word concepts, whereas MayoSRS has only 47\% and UMNSRS-Rel 0\%.
Representing multi-word concepts with word embeddings is likely to induce noise, whereas representing single-word concepts does not.

The Human Upper Bound (HUB), i.e.~the maximum Spearman's correlation achieved by any annotator with the mean rating, is 0.88. Note that this is a slightly biased metric as the mean rating will include the annotator's rating.
If the HUB is computed by comparing an annotator's rating with the mean rating of the other annotators, it is slightly lower at 0.70.
The HUB shows that there is large scope for improvement of relatedness models, but that it should not be expected to achieve performance scores of 0.9 or higher.

Our initial experiments show that \relB is a \emph{challenging new benchmark} for the performance analysis of concept relatedness models.

\section{Conclusions}
\label{sec:conclusions}
We presented a novel biomedical concept relatedness dataset sampled from EHR data, thus ensuring its relevance to EHR retrieval tasks. It is six times bigger than existing datasets, has high quality annotations, and complements existing datasets in terms of concept coverage.
Initial experiments showed that it is a challenging new benchmark for state-of-the-art biomedical word embedding models.

Despite our benchmark being much larger than existing datasets, we hope that this work inspires others to use our methodology for building even larger ones. As explained, our dataset covers around 3000 unique concepts, whereas SNOMED-CT consists of close to 350,000 concepts. We here focused on the most frequently co-occurring concept pairs. It would be interesting to expand this to less frequent pairs in future work. This could also involve focusing on specific areas of medicine.

Since our new benchmark consists of concept pairs expressed as 1) biomedical terms, 2) SNOMED IDs, and 3) UMLS CUIs, it can be used as a test bed for a large variety of concept representation models. In our initial experiments, we only considered word embedding models based on terms. In future work, it will be interesting to evaluate UMLS concept embeddings \cite{YuEtAl2017,BeamEtAl2018,HenryMM2019,ParkKHL2019} as well as graph-embeddings \cite{ChrichtonGPK2018,agarwal2019snomed2vec}.

Our work was motivated by the retrieval of information in EHRs related to a patient's presenting complaint. However, the usage of this benchmark goes far beyond this motivation. Coding in EHRs is not always perfect. For example, doctors do not always code both symptoms and diagnosis. Enabling the search for \emph{related} information is thus crucial to overcome the challenges associated with missing data.

\section*{Acknowledgements}
  We would like to thank all annotators for their help in constructing this new benchmark.

\bibliographystyle{coling}
\bibliography{references}

\begin{thebibliography}{}

\bibitem[\protect\citename{Agarwal \bgroup et al.\egroup
  }2019]{agarwal2019snomed2vec}
Khushbu Agarwal, Tome Eftimov, Raghavendra Addanki, Sutanay Choudhury, Suzanne
  Tamang, and Robert Rallo.
\newblock 2019.
\newblock Snomed2vec: Random walk and poincaré embeddings of a clinical
  knowledge base for healthcare analytics.
\newblock In {\em Proceedings of the 2019 KDD Workshop on Applied Data Science
  for Healthcare (DSHealth'19)}.

\bibitem[\protect\citename{Beam \bgroup et al.\egroup }2018]{BeamEtAl2018}
Andrew~L. Beam, Benjamin Kompa, Inbar Fried, Nathan Palmer, Xu~Shi, Tianxi Cai,
  and Isaac~S. Kohane.
\newblock 2018.
\newblock {Clinical Concept Embeddings Learned from Massive Sources of
  Multimodal Medical Data}.
\newblock {\em CoRR}.

\bibitem[\protect\citename{Bodenreider}2004]{Bodenreider_2004}
Olivier Bodenreider.
\newblock 2004.
\newblock {The Unified Medical Language System ({UMLS}): integrating biomedical
  terminology}.
\newblock {\em Nucleic Acids Research}, 32(90001):D267--D270.

\bibitem[\protect\citename{Chen \bgroup et al.\egroup }2019]{ChenPL2019}
Qingyu Chen, Yifan Peng, and Zhiyong Lu.
\newblock 2019.
\newblock Biosentvec: creating sentence embeddings for biomedical texts.
\newblock In {\em Proceedings of the 2019 IEEE International Conference on
  Healthcare Informatics (ICHI)}, pages 1--5.

\bibitem[\protect\citename{Chiu \bgroup et al.\egroup }2016]{ChiuEtAl2016}
Billy Chiu, Gamal Crichton, Anna Korhonen, and Sampo Pyysalo.
\newblock 2016.
\newblock How to train good word embeddings for biomedical {NLP}.
\newblock In {\em Proceedings of the 15th Workshop on Biomedical Natural
  Language Processing (BioNLP'16)}, pages 166--174.

\bibitem[\protect\citename{Chiu \bgroup et al.\egroup }2018]{Chiu2018}
Billy Chiu, Sampo Pyysalo, Ivan Vuli{\'{c}}, and Anna Korhonen.
\newblock 2018.
\newblock {Bio-SimVerb and Bio-SimLex: Wide-coverage evaluation sets of word
  similarity in biomedicine}.
\newblock {\em BMC Bioinformatics}, 19(33):1--13.

\bibitem[\protect\citename{Cicchetti}1994]{Cicchetti94}
Domenic Cicchetti.
\newblock 1994.
\newblock {Guidelines, Criteria, and Rules of Thumb for Evaluating Normed and
  Standardized Assessment Instrument in Psychology}.
\newblock {\em Psychological Assessment}, 6:284--290.

\bibitem[\protect\citename{Crichton \bgroup et al.\egroup
  }2018]{ChrichtonGPK2018}
Gamal Crichton, Yufan Guo, Sampo Pyysalo, and Anna Korhonen.
\newblock 2018.
\newblock {Neural networks for link prediction in realistic biomedical graphs:
  A multi-dimensional evaluation of graph embedding-based approaches}.
\newblock {\em BMC Bioinformatics}, 19(176):1--11.

\bibitem[\protect\citename{Donnelly}2006]{Donnelly2006}
Kevin Donnelly.
\newblock 2006.
\newblock {SNOMED-CT}: The advanced terminology and coding system for
  {eHealth}.
\newblock {\em Studies in Health Technology and Informatics}, 121:279.

\bibitem[\protect\citename{Flaxman}2015]{Flaxman2015}
Penny Flaxman.
\newblock 2015.
\newblock The 10-minute appointment.
\newblock {\em British Journal of General Practice}, 65(640):573--574.

\bibitem[\protect\citename{Henry \bgroup et al.\egroup }2018]{Henry2018}
Sam Henry, Clint Cuffy, and Bridget~T. McInnes.
\newblock 2018.
\newblock {Vector representations of multi-word terms for semantic
  relatedness}.
\newblock {\em Journal of Biomedical Informatics}, 77:111--119.

\bibitem[\protect\citename{Henry \bgroup et al.\egroup }2019]{HenryMM2019}
Sam Henry, Alex McQuilkin, and Bridget~T. McInnes.
\newblock 2019.
\newblock {Association measures for estimating semantic similarity and
  relatedness between biomedical concepts}.
\newblock {\em Artificial Intelligence in Medicine}, 93:1--10.

\bibitem[\protect\citename{Hliaoutakis}2005]{hliaoutakis2005semantic}
Angelos Hliaoutakis.
\newblock 2005.
\newblock {\em Semantic similarity measures in MeSH ontology and their
  application to information retrieval on Medline}.
\newblock Master’s thesis, Techical University of Crete.

\bibitem[\protect\citename{Kosmopoulos \bgroup et al.\egroup
  }2016]{KosmopoulosAP2015}
Aris Kosmopoulos, Ion Androutsopoulos, and Georgios Paliouras.
\newblock 2016.
\newblock {Biomedical Semantic Indexing using Dense Word Vectors in BioASQ}.

\bibitem[\protect\citename{McDonald \bgroup et al.\egroup
  }2018]{McdonaldEtAl2018}
Ryan McDonald, George Brokos, and Ion Androutsopoulos.
\newblock 2018.
\newblock Deep relevance ranking using enhanced document-query interactions.
\newblock In {\em Proceedings of the 2018 Conference on Empirical Methods in
  Natural Language Processing (EMNLP'18)}, pages 1849--1860.

\bibitem[\protect\citename{McGraw and Wong}1996]{Mcgraw96}
Kenneth~O. McGraw and S.P. Wong.
\newblock 1996.
\newblock {Forming Inferences About Some Intraclass Correlation Coefficients}.
\newblock {\em Psychological Methods}, 1(1):30--46.

\bibitem[\protect\citename{McInnes \bgroup et al.\egroup }2009]{McInnesPP2009}
Bridget~T. McInnes, Ted Pedersen, and Serguei~V.S. Pakhomov.
\newblock 2009.
\newblock {UMLS-Interface and UMLS-Similarity: open source software for
  measuring paths and semantic similarity}.
\newblock In {\em Proceedings of the Annual {AMIA} Symposium (AMIA'09)}, pages
  431--435.

\bibitem[\protect\citename{Morrison \bgroup et al.\egroup
  }2014]{MorrisonEtAl2014}
Zoe Morrison, Bernard Fernando, Dipak Kalra, Kathrin Cresswell, and Aziz
  Sheikh.
\newblock 2014.
\newblock National evaluation of the benefits and risks of greater structuring
  and coding of the electronic health record: exploratory qualitative
  investigation.
\newblock {\em Journal of the American Medical Informatics Association},
  21:492--500.

\bibitem[\protect\citename{Pakhomov \bgroup et al.\egroup }2010]{Pakhomov2010}
Serguei Pakhomov, Bridget McInnes, Terrence Adam, Ying Liu, Ted Pedersen, and
  Genevieve~B. Melton.
\newblock 2010.
\newblock {Semantic Similarity and Relatedness between Clinical Terms: An
  Experimental Study}.
\newblock In {\em Proceedings of the Annual AMIA Symposium (AMIA'10)}, pages
  572--576.

\bibitem[\protect\citename{Pakhomov \bgroup et al.\egroup }2011]{Pakhomov2011}
Serguei~V.S. Pakhomov, Ted Pedersen, Bridget McInnes, Genevieve~B. Melton,
  Alexander Ruggieri, and Christopher~G. Chute.
\newblock 2011.
\newblock {Towards a framework for developing semantic relatedness reference
  standards}.
\newblock {\em Journal of Biomedical Informatics}, 44(2):251--265.

\bibitem[\protect\citename{Park \bgroup et al.\egroup }2019]{ParkKHL2019}
Junseok Park, Kwangmin Kim, Woochang Hwang, and Doheon Lee.
\newblock 2019.
\newblock {Concept embedding to measure semantic relatedness for biomedical
  information ontologies}.
\newblock {\em Journal of Biomedical Informatics}, 94:103182.

\bibitem[\protect\citename{Pedersen \bgroup et al.\egroup }2007]{Pedersen2007}
Ted Pedersen, Serguei~V.S. Pakhomov, Siddharth Patwardhan, and Christopher~G.
  Chute.
\newblock 2007.
\newblock {Measures of semantic similarity and relatedness in the biomedical
  domain}.
\newblock {\em Journal of Biomedical Informatics}, 40(3):288--299.

\bibitem[\protect\citename{Pivovarov and Elhadad}2012]{Pivovarov2012}
Rimma Pivovarov and No{\'{e}}mie Elhadad.
\newblock 2012.
\newblock {A hybrid knowledge-based and data-driven approach to identifying
  semantically similar concepts}.
\newblock {\em Journal of Biomedical Informatics}, 45(3):471--481.

\bibitem[\protect\citename{Pyysalo \bgroup et al.\egroup
  }2013]{PyysaloEtAl2013}
Sampo Pyysalo, Filip Ginter, Hans Moen, Tapio Salakoski, and Sophia Ananiadou.
\newblock 2013.
\newblock {Distributional Semantics Resources for Biomedical Text Processing}.
\newblock In {\em Proceedings of the 5th Languages in Biology and Medicine
  Conference (LBM'13)}, pages 39--44.

\bibitem[\protect\citename{Robinson \bgroup et al.\egroup
  }1997]{RobinsonEtAl2997}
David Robinson, Erich Schulz, Philip Brown, and Colin Price.
\newblock 1997.
\newblock {Updating the Read Codes: User-interactive Maintenance of a Dynamic
  Clinical Vocabulary}.
\newblock {\em Journal of the American Medical Informatics Association},
  4(6):465--472.

\bibitem[\protect\citename{Salisbury}2019]{Salisbury2019}
Helen Salisbury.
\newblock 2019.
\newblock Helen salisbury: The 10 minute appointment.
\newblock {\em BMJ}, 365.

\bibitem[\protect\citename{Schulz and Juric}2020]{SchulzJ2020}
Claudia Schulz and Damir Juric.
\newblock 2020.
\newblock Can embeddings adequately represent medical terminology? new
  large-scale medical term similarity datasets have the answer!
\newblock In {\em Proceedings of the 34th {AAAI} Conference on Artificial
  Intelligence (AAAI'20)}, pages 8775--8782.

\bibitem[\protect\citename{Sinha \bgroup et al.\egroup }2012]{ehrBook}
Pradeep Sinha, Gaur Sunder, Prashant Bendale, Manisha Mantri, and Atreya Dande.
\newblock 2012.
\newblock {\em Electronic Health Record: Standards, Coding Systems, Frameworks,
  and Infrastructures}.
\newblock Wiley-IEEE Press.

\bibitem[\protect\citename{Smalheiser \bgroup et al.\egroup
  }2019]{SmalheiserCB2019}
Neil~R. Smalheiser, Aaron~M. Cohen, and Gary Bonifield.
\newblock 2019.
\newblock {Unsupervised low-dimensional vector representations for words,
  phrases and text that are transparent, scalable, and produce similarity
  metrics that are not redundant with neural embeddings}.
\newblock {\em Journal of Biomedical Informatics}, 90:103096.

\bibitem[\protect\citename{Yu \bgroup et al.\egroup }2017]{YuEtAl2017}
Zhiguo Yu, Byron~C. Wallace, Todd Johnson, and Trevor Cohen.
\newblock 2017.
\newblock {Retrofitting Concept Vector Representations of Medical Concepts to
  Improve Estimates of Semantic Similarity and Relatedness}.
\newblock {\em Studies in Health Technology and Informatics}, 245:657--661.

\bibitem[\protect\citename{Zhang \bgroup et al.\egroup }2019]{ZhangEtAl2019}
Yijia Zhang, Qingyu Chen, Zhihao Yang, Hongfei Lin, and Zhiyong Lu.
\newblock 2019.
\newblock {BioWordVec, improving biomedical word embeddings with subword
  information and MeSH}.
\newblock {\em Scientific Data}, 6(1):52.

\bibitem[\protect\citename{Zhelezniak \bgroup et al.\egroup
  }2019]{ZhelezniakEtAl2019-fuzzyJaccard}
Vitalii Zhelezniak, Aleksandar Savkov, April Shen, Francesco Moramarco, Jack
  Flann, and Nils Hammerla.
\newblock 2019.
\newblock Don't settle for average, go for the max: Fuzzy sets and max-pooled
  word vectors.
\newblock In {\em Proceedings of the 7th International Conference on Learning
  Representations (ICLR'19)}.

\end{thebibliography}

\end{document}